\documentclass[journal]{IEEEtran}
\usepackage[utf8]{inputenc} 
\usepackage[T1]{fontenc}    
\usepackage{cite}
\usepackage{amsmath,amssymb,amsfonts}
\usepackage{mathtools}
\usepackage{url,hyperref,lineno,microtype,subcaption}
\usepackage{graphicx}
\usepackage{xcolor}
\usepackage{multirow}
\usepackage{hyperref}
\usepackage{algorithm}
\usepackage[noend]{algpseudocode}
\usepackage{etoolbox}
\usepackage{mathtools}

\makeatletter
\newcommand*{\algrule}[1][\algorithmicindent]{%
  \makebox[#1][l]{%
    \hspace*{.2em}
  }
}

\newcount\ALG@printindent@tempcnta
\def\ALG@printindent{%
    \ifnum \theALG@nested>0
    \ifx\ALG@text\ALG@x@notext
    \else
    \unskip
    \ALG@printindent@tempcnta=1
    \loop
    \algrule[\csname ALG@ind@\the\ALG@printindent@tempcnta\endcsname]%
    \advance \ALG@printindent@tempcnta 1
    \ifnum \ALG@printindent@tempcnta<\numexpr\theALG@nested+1\relax
    \repeat
    \fi
    \fi
}
\patchcmd{\ALG@doentity}{\noindent\hskip\ALG@tlm}{\ALG@printindent}{}{\errmessage{failed to patch}}
\patchcmd{\ALG@doentity}{\item[]\nointerlineskip}{}{}{} 
\makeatother

\algnewcommand\algorithmicforeach{\textbf{for each}}
\algdef{S}[FOR]{ForEach}[1]{\algorithmicforeach\ #1\ \algorithmicdo}

\renewcommand{\algorithmicrequire}{\textbf{Input:}}
\renewcommand{\algorithmicensure}{\textbf{Output:}}

\usepackage{amsthm}
\theoremstyle{definition}
\newtheorem{definition}{Definition}[section]

\begin{document}

\title{Intrinsically
Motivated Hierarchical Policy Learning in Multi-objective Markov Decision Processes}

\author{Sherif~Abdelfattah,~\IEEEmembership{Member,~IEEE,}
        Kathryn~Merrick,~\IEEEmembership{Senior Member,~IEEE,}
        and~Jiankun~Hu,~\IEEEmembership{Senior~Member,~IEEE}}

\markboth{Journal of \LaTeX\ Class Files,~Vol.~14, No.~8, August~2015}%
{Shell \MakeLowercase{\textit{et al.}}: Bare Demo of IEEEtran.cls for IEEE Communications Society Journals}

\maketitle

\begin{abstract}
Multi-objective Markov decision processes are sequential decision-making problems that involve multiple conflicting reward functions that cannot be optimized simultaneously without a compromise. This type of problems cannot be solved by a single optimal policy as in the conventional case. Alternatively, multi-objective reinforcement learning methods evolve a coverage set of optimal policies that can satisfy all possible preferences in solving the problem. However, many of these methods cannot generalize their coverage sets to work in non-stationary environments. In these environments, the parameters of the state transition and reward distribution vary over time. This limitation results in significant performance degradation for the evolved policies sets. In order to overcome this limitation, there is a need to learn a generic skill set that can bootstrap the evolution of the policy coverage set for each shift in the environment dynamics therefore, it can facilitate a continuous learning process. In this work, intrinsically motivated reinforcement learning has been successfully deployed to evolve generic skill sets for learning hierarchical policies to solve multi-objective Markov decision processes. We propose a novel dual-phase intrinsically motivated reinforcement learning method to address this limitation. In the first phase, a generic set of skills is learned. While in the second phase, this set is used to bootstrap policy coverage sets for each shift in the environment dynamics. We show experimentally that the proposed method significantly outperforms state-of-the-art multi-objective reinforcement methods on a dynamic robotics environment.
 
\end{abstract}


\IEEEpeerreviewmaketitle

\section{Introduction}
Interactive learning is a vital human ability needed to acquire new skills. This is achieved through direct interaction with the environment that includes repetitive cycles of sense, think, and act. Similarly, reinforcement learning (RL) is a machine learning paradigm that aims at mimicking this human interactive learning ability. In a RL scenario, the learning agent evolves an optimal policy to achieve a specific objective by sensing its current state, performing an action, and observing a reward value (i.e., positive or negative) \cite{sutton1998}. The reward value is generated by a reward function that maps the state and action pair into a quantitative value that reflects how good or bad this pair is with respect to an objective. The objective is meant to guide the learning process (e.g., maximize overall score in a game) and it is coupled with a reward function.

Advances in deep learning techniques achieved over the last decade \cite{lecun2015deep} contributed to the enhancement of RL methods. Recent RL methods achieved noticeable breakthroughs such as playing Atari games with human-level performance \cite{mnih2015human} or beating the world champion in the game of Go \cite{silver2016mastering}. Many of these RL advances follow a single objective problem in which the agent learns guided by a single reward function. Albeit, there is a type of sequential decision-making problem that does not follow this assumption. In this type, there are multiple objectives that are naturally in conflict with each other and cannot be optimized simultaneously without a compromise. Consider a scenario for a search and rescue robot that aims at maximizing the number of rescued victims, minimizing exposure to risk in the environment (i.e., fire or flood) to prevent physical destruction, and minimizing the overall time to finish the job.  Another scenario can be a portfolio management bot that aims to maximize the investment return over different stock sectors (i.e., industrial, agricultural, education, etc.), each of these sectors will have a dedicated objective and they cannot be maximized simultaneously with limited funds. This type of problems poses a significant challenge on single objective RL methods as there is no single policy that can satisfy all the preferences to solve the problem.

Alternatively, multi-objective reinforcement learning (MORL) methods can deal with this type of problem by targeting a coverage set of policies instead of a single policy in the conventional RL case. Usually, this is achieved by exploring the preference space over the defined objectives and evolving an optimal policy for each legitimate preference \cite{Vamplew2011}. After training, the learning agent can switch between policies in the evolved coverage set to cope with different preferences. The difference between RL and MORL lies in the reward signal, which is a scalar value in the RL scenario and a vector in the MORL scenario.

Despite the effectiveness of current MORL methods, the majority of them assume that environment dynamics follow a stationary distribution \cite{roijers2013survey}. According to this assumption, the dynamics such as state transition probability distribution (i.e., types and order to entities in the environment) or reward probability distribution (i.e. types and numbers of objectives) are assumed to remain fixed between the training and the operation processes of the learning agent. However, this is not a realistic assumption in many real-world scenarios. For example, assuming an unmanned ground vehicle (UGV) in a playground (i.e., flat terrestrial surface) with obstacles, we may have different entities introduced to this setup over time such as entities of interest (i.e., treasures or victims) to detect or danger entities (i.e., fire or enemy)  to avoid. The introduction of new or different entities will also impact the objective space over the problem. If the learning method cannot cope efficiently with such dynamic variations, it will result in performance degradation and demand for re-initializing the training process from scratch after each variation. With the assumption that the agent's setup (i.e., UGV build) and the basic environment's characteristics (i.e., type of the terrestrial surface and common entities such as obstacles) remain the same between the dynamics variations, there is room for enhancement if we can use knowledge about stationary parts of the environment or the agent.

In a previous work \cite{Sherif_18}, we proposed a novel MORL method that can adapt effectively to non-stationary dynamics in state transition distribution given fixed types of entities and objectives in the problem. This was achieved through the concept of policy bootstrapping in which the preference space is decomposed into distinct regions and for each of them, we evolve one steppingstone policy. Steppingstone policies are used to bootstrap specialized policies in an online manner to adapt to different changes in the state transition distribution. In addition, we used an intrinsically motivated technique for adaptive preference exploration. However, this method cannot generalize to different types of entities and objectives without a significant update to its policy coverage set. 

In this paper, we propose a novel MORL method that can effectively generalize to non-stationary dynamics in the state transition distribution and reward distribution.  This is achieved over two learning phases. In the first phase, we evolve a generic set of skills based on stationary aspects of the problem including the learning agent's build and the stationary characteristics of the environment. While in the second phase, we use this skill set to bootstrap the policy coverage set learning process to deal with shifts in the non-stationary dynamics.  During the skill learning phase, we use a competency-based intrinsically motivated reinforcement learning (IMRL) method \cite{Oudeyer2007} in order to developmentally learn skills that match the current skill level of the learning agent. Therefore, we can enhance the performance of learned skills. During the policy coverage set learning phase, we extend our previous work \cite{Sherif_18} to learn hierarchical policies using the generic skill set. We show experimentally the effectiveness of the proposed method in generalizing over three different robotic scenarios compared to state-of-the-art MORL methods.

The main contribution of this paper can be summarized as follows:
\begin{itemize}
\item We propose an effective technique to learn skill sets in a developmental manner using intrinsic motivation reinforcement learning.
\item We propose a novel multi-objective reinforcement learning method for evolving policy coverage sets using hierarchical policy learning.
\item We experimentally evaluate the generalization over three robotic scenarios and compare with state-of-the-art methods.
\end{itemize}

The remainder of the paper is organized as follows. Section \ref{sec:Background} introduces background concepts and defines the research problem. Section \ref{sec:RelatedWork} overviews related work. Section \ref{sec:Methodology} presents the proposed methodology. Section \ref{sec:ExperimentalDesign} explains the experimental design. Section \ref{sec:Results} shows and discusses the experimental results. Finally, Section \ref{sec:conc} concludes the work and indicates possible future extensions. 

\section{Background} 
\label{sec:Background}
This section introduces the fundamental concepts and formulates the research problem.
\subsection{Markov Decision Processes (MDPs)}
A Markov decision process (MDP) is a sequential planning problem in which the learning agent senses its current state in the environment ($s_t$) at time $t$, performs an action ($a_t$) which results in transition to a new state ($s_{t+1}$) and receives a reward/penalty ($r_{t+1}$) for reaching this new state \cite{Tsitsiklis87}.
The MDP can be represented with a tuple $\left\langle S,A,\mathbb{P_{\mathrm{ss^{'}}}},R\,:\,s_{t},a_{t}\rightarrow r_{t+1},\mu,\gamma\right\rangle $. Where $S$ is the state space, $A$ is the action space, $\mathbb{P_{\mathrm{ss^{'}}}\mathrm{=Pr(\mathit{s_{t+1}}=\mathit{s^{'}}|\mathit{s_{t}}=\mathit{s,a_{t}}=\mathit{a};\theta_{s})}}$ is the state transition probability distribution, $R\,:\,s_{t},a_{t}\rightarrow r_{t+1}$ is the reward function that maps a state and action pair into a reward value, $\mu$ is the initial state probability distribution, and $\gamma$ is the discounting factor for balancing the bias between immediate and future rewards. 
Usually the aim of the learning agent is to find a policy $\pi^{*}\,:\,s_{t}\rightarrow a_{t}$ that maximizes an aggregation function over rewards, typically the discounted sum of rewards over the time horizon $T$:

\begin{equation}
\sum_{t=1}^{T}\gamma_{t}r_{t}
\end{equation}

\subsection{Multi-objective Markov Decision Processes (MOMDPs)}
The multi-objective Markov decision process (MOMDP) extends the MDP problem by allowing more than one reward function to exist, each representing a unique objective. Accordingly, the MOMDP problem can be represented by the tuple $\ensuremath{\left\langle S,A,\mathbb{P_{\mathrm{ss^{'}}}},\vec{r},\mu,\gamma\right\rangle }$, where $\vec{r}=[r^1,r^2,\ldots,r^N]$ is a vector of rewards dedicated to $N$ objectives. Multiple rewards can be combined using a user's preference $P$ and a scalarization function $f^{\dagger}$.

\theoremstyle{definition}
\begin{definition}{Preference:}
\label{def:Pref}
A user's preference $\ensuremath{P=[w^{1},w^{2},\ldots,w^{N}]}\,\forall w^{n}\in\{0,1\}$ represents a specific prioritization across the observed reward vector $\vec{R}$. This preference is constrained to have its element-wise sum equals to one $\sum_{n=1}^{N}w^{n}=1$.
\end{definition}

\theoremstyle{definition}
\begin{definition}{Scalarization Function:}
\label{def:ScalFunc}
A scalarization function combines an observed vector of reward into a scalar reward value given a user's preferences.
\begin{center}
$f^{\dagger}:\,\vec{r},P\rightarrow r^{\dagger}$
\end{center} 
\end{definition}

\theoremstyle{definition}
\begin{definition}{Policy:}
 A policy ($\pi$) is a mapping from a state space $S$ to an action space $A$ that achieves a task-specific objectives.
\end{definition}

\theoremstyle{definition}
\begin{definition}{Skill:}
 A skill ($\eta$) is a mapping from a state space $S$ to an action space $A$ that achieves a generic objective.
\end{definition}

The aim of the learning agent is to find a policy coverage set $\Pi^{*}$ that contains an optimal policy for any given user's preference $P^i$ such that $\arg\max_{\pi\in\Pi}(\sum_{t=1}^{T}\gamma_{t}r_{t}^{\dagger i})\in\Pi^{*}$, where $\Pi$ is the policy search space.

\subsection{Problem Definition}
The research problem in this work is two fold. First, we need to learn a generic skill set independently from any specific task. Given a goal generation function $f\,:\,X^{'},S\rightarrow\vec{x}_{t+k}$
, that uses the set of previously generated goals $X^{'}$, and
the state space of the environment (including the internal state of
the agent) $S$, to generate a vector of goals $\vec{x}_{t+k}$ to
be optimized in the time period $k$. We need
to find the skill set $U$ that maximizes the reward return of the
learning agent over the set of all generated goals $G$ during the learning
time horizon $T$ and over $E$ total time steps assigned for learning each goal as follows:

\begin{equation}
\max\sum_{t=0}^{T}\sum_{g=1}^{X}\sum_{e=0}^{E}\gamma^{e}r_{t+e+1}^{x}
\end{equation}

The second part of the problem is to deal with shifts in the environment dynamics including the parameters of the state transition probability distribution $\theta_{S}$ and the parameters of the rewards prior distribution $\mathbb{P}_{R}=Pr(r^{n};\theta_{r})\,\forall r^{n}\in\{r^{1},r^{2},\ldots,r^{N}\}$. The aim is given the set of generic skills $U$, to find the optimum policy coverage set $\Pi^{*}$ that can satisfy any user's preference under the current dynamics parameterization. The optimum policy coverage set $\Pi^{*}$ has to maximize the scalarized reward return for any given set of preferences within a $T$ \textit{time horizon}:
\begin{align}
\begin{split}
\max\;R_{t}^{\vec{w}^{i}} & =\sum_{t=0}^{T}\gamma^{t}f^{\dagger}\,(\vec{r}_{t},P^{i})\\
\end{split}
\\
\begin{split}
 & s.t.\:P^{i}\in W\,\forall\,P^{i}\in\mathbb{R^{\mathit{M}}\mathit{,\sum_{m=1}^{M}w^{m}}=\mathit{1}}\nonumber 
\end{split}
\end{align}

\noindent where $W$ is the set of all legitimate user's preferences over the defined objectives.

\section{Related Work}
\label{sec:RelatedWork}
In this section, we review the related work in intrinsically motivated skill learning and multi-objective reinforcement learning literature in order to contrast the contribution of our paper. 

\subsection{Intrinsically Motivated Skill Learning Methods}

Learning skill sets through intrinsic motivation has been explored previously in the literature for solving single objective reinforcement learning tasks through hierarchical policy learning. A common theme around these methods is that they usually include two key components that communicate with each other in order to evolve the skill set. These two components can be generally described as proposer and learner. The proposer is meant to explore the goal space with the objective to sample new goals that can enhance the current performance level of the learner, while the learner will aim at responding to each newly proposed goal with an optimum policy that reflects its skill in performing this goal. Many examples of this theme can be found in the literature. An adversarial game-play was proposed by Schmidhuber \cite{Schmidhuber2003} in which two predictive agents compete with each other by proposing and accepting prediction bets on state transitions. Similarly, a self-play strategy was proposed by Sukhbaatar et al. \cite{SukhbaatarKSF17} in which two minds (Alice and Bob) were competing with each other to learn new skills. Alice learns to propose goals that are not optimally covered by Bob’s current skill set, while Bob aims at evolving optimal policies to perform the proposed goals. 

A different approach is to allow extrinsic reward signals to guide the skill learning process. A coupled network architecture with two levels was proposed by Kulkarni et al. \cite{Kulkarni2016} to learn hierarchical policies. In the first level, there is a meta-controller that samples goals to maximize an extrinsic scenario-specific reward signal, while in the second level, there is a controller that is guided by an intrinsic reward signal to reach the sampled goals by sampling actions. Similarly, Dilokthanakul et al. \cite{Dilokthanakul17} proposed a two-level hierarchical policy learning framework that includes a meta-controller and a controller, with a difference of allowing the extrinsic reward signal to propagate to the controller via a trade-off parameter that balance the intrinsic and extrinsic reward effects. One limitation with this approach is that the learned skill set is biased towards a specific scenario as it uses the extrinsic reward during the skill learning process. This will demand to retrain for adapting to different scenarios.

Although this theme of learning skill sets through intrinsic motivation proved its effectiveness in solving MDP scenarios, it has not been explored yet in solving MOMDP scenarios to the best of our knowledge. In this paper, we adopt a competitive adversarial self-play theme in learning a generic skill set in isolation from any scenario-specific extrinsic reward during the skill learning phase, then we evolve hierarchical policies to solve multi-objective scenarios using this generic skill set in a later phase.

\subsection{Multi-objective Reinforcement Learning Methods}

Mainly there are two broad categories of MORL methods: single policy category; and multiple policy category \cite{roijers2013survey}. The former category assumes that the user’s preference is defined beforehand solving the MOMDP problem and uses either a scalarization function to convert the problem into a conventional MDP scenario \cite{Moffaert2013, Castelletti2013, lizotte2010efficient, Perny_2010, Ogryczak_2011}, or a constrained representation of the problem in which one reward function will be optimized while considering the other functions as constraints for the optimization \cite{ Feinberg_95,Altman_1999 }. One major limitation of this category is the difficulty of satisfying its main assumption of having a predefined user’s preference in many real-world scenarios. The latter category addresses this limitation by evolving a policy coverage set that can satisfy any user’s preference in solving the MOMDP problem. Basically, this is achieved through two main components: the preference explorer; and the policy optimizer. The former component is responsible for sampling preferences from the user’s preference space in order to evolve a diverse policy coverage set. While the latter component aims at evolving optimal policies that can effectively satisfy the sampled preferences. Methods in this category differ in the preference exploration mechanism such as evolutionary exploration \cite{Busa-Fekete2014}, threshold lexicographic ordering (TLO) \cite{gabor1998multi}, or optimistic linear support (OLS) \cite{Roijers2014}. 

It can be noticed that the multiple policy approaches are symmetric to the self-play theme in intrinsically motivated methods, which we previously exploited to propose a novel intrinsically motivated method to evolve the policy coverage set \cite{Sherif_18}. In this paper, we extend this method to generalize over different scenarios through hierarchical policy learning.

\section{Methodology}
\label{sec:Methodology}
The working mechanism of the proposed methodology has two separate stages. First, we start with the generic skill learning stage in which we learn a set of generic skills that can be used with different task scenarios given the learning agent's build and the stationary characteristics of the environment. The main assumption behind this stage is that such generic skill set can be reused to evolve specialized policies for different task scenarios (i.e., objectives and environment layout). As our aim is to learn hierarchical policy coverage sets in MOMDPs based on generic skill sets, we assume that the target set of generic skills is already defined in the environment.

\begin{figure}
\begin{centering}
\includegraphics[width=9cm, height=11cm]{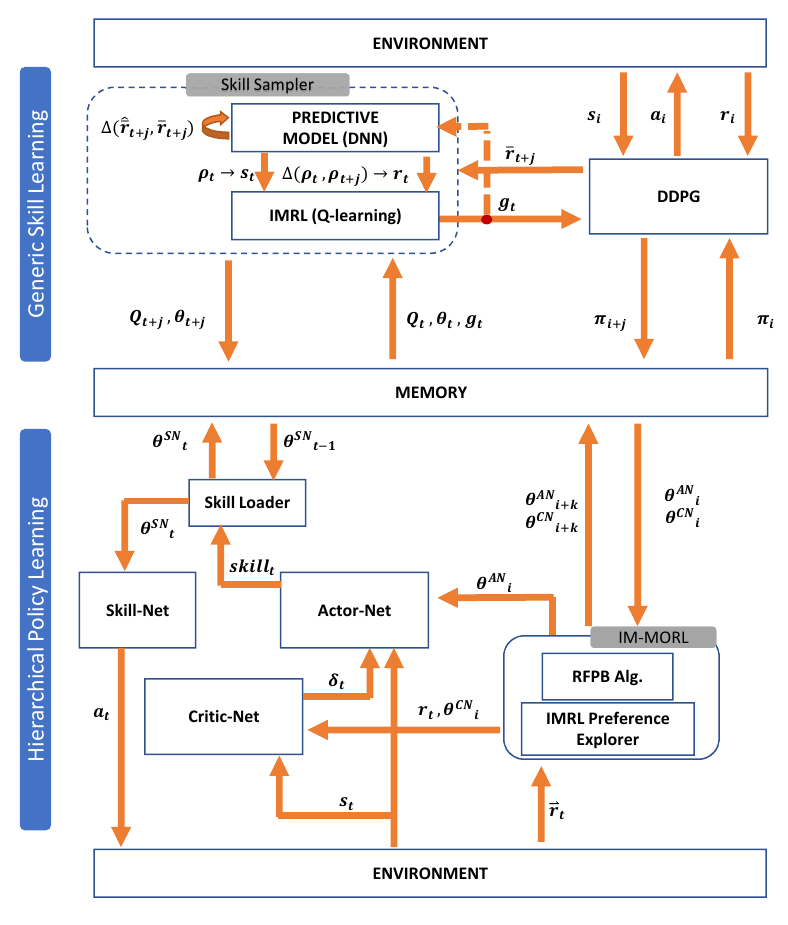}
\par\end{centering}
\caption{A block diagram for the design of the proposed model. The generic skill learning stage is shown in the top section, while the hierarchical policy learning stage is presented in the bottom section.}
\label{fig:comb_model}

\end{figure}

The top section in Figure \ref{fig:comb_model} depicts a block diagram for learning the generic skill set. There are two main building blocks in this design: the skill sampler, and the skill optimizer. The former aims at sampling skills from the predefined skill set that matches the current performance level of the optimizer (i.e., not too easy or too hard). This is achieved through IMRL. We used a competency-based IMRL design~\cite{Oudeyer_2007} for the skill sampler component, which has been followed in the state-of-the-art skill learning methods \cite{SukhbaatarKSF17,Dilokthanakul17}. In this IMRL paradigm, a predictive model (implemented as a deep feed-forward neural network) tries to predict the average reward value for the sampled skill achieved by the skill optimizer. While a reinforcement learning algorithm learns which skill to sample given a state representing the current performance level of the predictive model in terms of prediction accuracy ($\rho_{t}$) and an intrinsically generated reward signal formulated as the performance progress of the predictive model before and after sampling the latest skill ($\Delta(\rho_{t},\rho_{t+j})$). Table \ref{tbl:DNN} lists the neural network design configuration of the predictive model. We used the Q-learning \cite{Watkins1992} reinforcement learning algorithm for the IMRL component. For the skill optimizer, we use the same actor-critic architecture design of the original deep deterministic policy gradient (DDPG) algorithm \cite{DDPG2016}, while changing the state representation layers to cope with our learning agent's build. 

\setlength{\textfloatsep}{2.5pt}

\bgroup
\def\arraystretch{1.3}
\begin{table}

\caption{Design configuration of the skill sampler's predictive model as a feed-forward neural network }

\begin{centering}
\begin{tabular}{c c}
\hline 
Configuration & Value\tabularnewline
\hline 
\hline 
Layers & Linear(10),ReLU(32),ReLU(16),tanh(8),Linear(1) \tabularnewline
$\alpha$ & $0.085$ \tabularnewline
Dropout & $0.35$ \tabularnewline
Cost Function & cross entropy \tabularnewline
Optimizer & ADAM \tabularnewline
\hline 
\end{tabular}
\par\end{centering}
\label{tbl:DNN}
\end{table}
\egroup

After finishing the first stage, the agent starts the second stage which builds over the learned skill set to evolve a coverage set of policies that can solve the MOMDP problem. We reuse our previously proposed intrinsically motivated multi-objective reinforcement learning IM-MORL method \cite{Sherif_18} for evolving robust policy coverage set in MOMDP problems. The IM-MORL method works through two main components that learn in an adversarial manner. The first component is the preference explorer component which includes an IMRL algorithm to sample preference that the current policy coverage set $\Pi^{*}$ is not performing well on them. This is achieved by maximizing the enhancement in the prediction performance of a neural network that targets predicting the performance of $\Pi^{*}$ for the sampled preference. Therefore, the IMRL tends to sample preferences that are challenging to predict due to instability in the $\Pi^{*}$ performance. While the second component is the robust fuzzy policy bootstrapping (RFPB) algorithm (see Algorithm \ref{alg:RFPB}), which is a MORL algorithm that takes preference proposed by the former component to evolve the policy coverage set $\Pi^{*}$. The RFPB algorithm evolves robust policy coverage sets through the concept of policy bootstrapping \cite{Sherif_18}.

The RFPB algorithm follows a policy bootstrapping strategy that divides the user's preference space over the defined objectives into a finite number of regions using a fuzzy representation of preferences. Thus, a preference region $g^{i}$ is a unique combination of fuzzy values for the weight components (e.g., the combination [low, high] in a two objective space scenario). The fuzzy membership (referred to as FuzzyMembership in Algorithm \ref{alg:RFPB}) is calculated through the triangular fuzzy membership function \cite{PEDRYCZ_199421}. For each region, the RFPB algorithm evolves an optimal steppingstone policy ($p$) that can bootstrap policies for different preferences that fall in that region. Following this strategy, the algorithm can reduce the search space for optimum policies and adapt more robustly in an online manner to dynamics in the environment. The evaluation of steppingstone policies is based on a robustness metric ($\beta$) that represents a trade-off between a policy's performance and its robustness to dynamics in the environment \cite{Sherif_18}. When the current preference region changes (i.e., the new user's preferences falls into another region), the RFPB algorithm saves the parameters of the current steppingstone policy into the policy coverage set $\Pi$ and loads the parameters of the dedicated policy of the new region if it is exists, otherwise, it initializes it from the best policy of adjacent regions. Finally, it passes the parameters ($\theta^{\prime}$) of the policy to the policy optimization algorithm (i.e., reinforcement learning algorithm) to optimize it given the new user's preference. Following this generic workflow, the RFPB algorithm can adopt any policy optimization algorithm that suits the characteristics of the problem.

For the policy optimization algorithm, we propose a variant of the deep deterministic policy gradient algorithm (DDPG) \cite{DDPG2016} that can evolve hierarchical policies (i.e., policies over skills instead of atomic actions). We name this algorithm as hierarchical deep deterministic policy gradient (HDDPG). Algorithm \ref{alg:HDDPG} presents the working steps of the HDDPG algorithm.

\begin{algorithm}
\caption{ Robust Fuzzy Policy Bootstrapping (RFPB) \cite{Sherif_18}}\label{alg:RFPB}
\begin{algorithmic}[1]
  \Require Preferences at times $t$ and $t-1$ ($P_{t},P_{t-1}$).
  \State Get the fuzzy region of the new preference $FuzzyMembership(P_{t})\rightarrow g^{i}$
  \If{$\pi(g^{i})\neq\emptyset$}
  \State $\pi^{'}\coloneqq \pi(g^{i})$
  \ElsIf{$\pi(g^{i-1})\neq\emptyset\;and\;\pi(g^{i+1})\neq\emptyset$}
  \State $\pi^{'}\coloneqq\arg\max_{\pi\in\{\pi(g^{i-1}),\,\pi(g^{i+1})\}}\beta(\pi)$
  \ElsIf{$\pi(g^{i-1})=\emptyset\;and\;\pi(g^{i+1})=\emptyset$}
  \State $\pi^{'}\coloneqq\phi$
  \Else 
  \State $\pi^{'}\coloneqq\arg_{\pi\in\{\pi(g^{i-1}),\,\pi(g^{i+1})\}}\pi\neq\phi$
  \EndIf
  \State Get the fuzzy region of the old preference $FuzzyMembership(P_{t-1})\rightarrow g^{j}$ 
  \If{$\pi(g^{j})\neq\emptyset$}
  \State $\pi(g^{j})\coloneqq\arg\max_{\pi\in\{\pi(g^{j}),\,\pi^{P_{t-1}}\}}\beta(\pi)$
  \Else
  \State $\pi(g^{j})\coloneqq\,\pi^{P_{t-1}}$
  \EndIf
  \State Store $\,\pi(g^{j})$ in $\Pi$ 
  \If{$\pi^{'}=\emptyset$}
  \State $\theta^{'}\coloneqq\phi$
  \Else
  \State $\theta^{'}\coloneqq\theta(\pi^{'})$
  \EndIf
  \State Follow the hierarchical deterministic deep policy gradient algorithm, HDDPG($P_{t},\,\theta^{\prime}$)
  \end{algorithmic}
\end{algorithm}

\begin{algorithm}
\caption{Hierarchical Deterministic Deep Policy Gradient (HDDPG)}\label{alg:HDDPG}
\begin{algorithmic}[1]
\Require User's preference $P_{t}$ and parameter set $\theta^{\prime}=\{\theta^{Q},\theta^{\mu}\}$.
\If{$\theta^{\prime}=\phi$}
  \State Randomly initialize critic $Q(s,\eta\,|\,\theta^{Q})$ and actor $\mu(s\,|\,\theta^{\mu})$ networks with parameters $\theta^{Q}$ and $\theta^{\mu}$
  \Else
  \State Initialize critic $Q(s,\eta\,|\,\theta^{Q})$ and actor $\mu(s\,|\,\theta^{\mu})$ networks with parameters from $\theta^{\prime}$
  \EndIf
\State Initialize target networks $Q^{\prime}$ and $\mu^{\prime}$ with parameters $\theta^{Q^{\prime}}\leftarrow\theta^{Q}$ and $\theta^{\mu^{\prime}}\leftarrow\theta^{\mu}$
\State Initialize reply buffer $R$
\For{episode = 1, $M$}
        \State Initialize an Ornstein-Uhlenbeck process $\mathcal{N}$ for skill exploration.
        \State Observe initial state $s_1$.
        \For{$t = 1$, $T$}
            \State Select skill $\eta_t=\mu(s_t\,|\,\theta^{\mu})+\mathcal{N}_t$
            \State Retrieve parameters $\theta^{\eta_t}$ of $\eta_t$ from skill set $U$
            \State Execute the skill network $\mu^{\eta_t}(s\,|\,\theta^{\eta_t})$
            \State Observer scalarized reward $r^{\dagger}_{t}=P_{t}\cdot\vec{r}_{t}$
            \State Observe new state $s_{t+1}$
            \State Save transition $(s_t,\eta_t,r^{\dagger}_{t},s_{t+1}) $ into $R$
            \State Sample a random batch of $V$ transitions \\    $(s_i,\eta_i,r^{\dagger}_{i},s_{i+1})$ from $R$ 
            \State Set $y_i=r^{\dagger}_{i}+\gamma\,Q^{\prime}(s_{i+1},\mu^{\prime}(s_{i+1}|\theta^{\mu^{\prime}})|\theta^{Q^{\prime}})$
            \State Update critic by minimizing the loss: $L= \frac{1}{V}\,\sum_{i=1}^{V}(y_{i}-Q(s_{i},\eta_{i}|\theta^{Q}))^{2}$
            \State Update the actor policy using the sampled policy gradient $\nabla_{\theta^{\mu}}J \thickapprox\frac{1}{V}\,\sum_{i=1}^{V}\nabla_{\eta}Q(s,\eta\,|\theta^{Q})\,|_{s=s_{i},\eta=\mu(s_{i})}\nabla_{\theta^{\mu}}\mu(s\,|\theta^{\mu})\,|_{s_{i}}$
            \State Update the target networks:
            \State $\theta^{Q^{\prime}\leftarrow}  \tau\theta^{Q}+(1-\tau)\theta^{Q^{\prime}}$
            \State $\theta^{\mu^{\prime}}\leftarrow  \tau\theta^{\mu}+(1-\tau)\theta^{\mu^{\prime}}$
        \EndFor
        \State \textbf{endfor}
\EndFor
\State \textbf{endfor}
\end{algorithmic}
\end{algorithm}
\setlength{\textfloatsep}{2.5pt}

  For further illustration of the design of this stage, the bottom section in Figure \ref{fig:comb_model} presents a block diagram that explains the underlying working mechanism. The RFPB algorithm takes care of bootstrapping the parameter configuration (i.e., parameters for actor and critic networks) of the HDDPG algorithm. The HDDPG algorithm consists of four components. First, the critic network (Critic-Net) which approximates the Q-value of the current state-skill pairs guided by the temporal difference error ($\delta$), see Equation \ref{eq:TD_Error}. The actor network (Actor-Net) learns to pick the right skill ($\eta$) for the current state using the policy gradient function ($\nabla_{\theta^{\mu}}J$ in Algorithm \ref{alg:HDDPG}) provided the generated temporal difference error ($\delta$) from the Critic-Net.  The actions exploration is performed using the Ornstein–Uhlenbeck stochastic process, that can produce temporally correlated exploration noise for smooth transitions across action values. This process is calculated as per Equation \ref{eq:OU}. After identifying the skill to be executed by the Actor-Net, the skill loader component retrieves the parameters ($\theta^{\eta}$) of the skill's learned policy in the previous stage from memory and assigned it to the action execution network (Skill-Net) to perform it. The Skill-Net works as an architecture template that is loaded with the previously learned parameters without any learning in this stage. The memory component in Figure \ref{fig:comb_model} mainly represents the policy coverage set $\Pi$ and the skill set $U$. 
  
\begin{equation}
da_{t}=\theta(\mu-a_{t})dt+\sigma dW_{t}
\label{eq:OU}
\end{equation}

\noindent where $\theta$,$\sigma$, and $\mu$ are parameters and $W_{t}$ represents the Wiener process, which is a stochastic process that is initialized as $W_{0}=0$ and at each following time step it is incremented by a Gaussian random value $(W_{t}-W_{s})\sim\mathcal{N}(0,t-s)\,\forall\, 0\leq s<t$.  

\begin{equation}
\delta_{t}=r^{\dagger}_{t+1}+\gamma\max_{\eta^{'}}Q\,(s_{t+1},\eta^{'})-Q\,(s_{t},\eta_{t})
\label{eq:TD_Error}
\end{equation}

It has to be noted that the transition between the first stage and the second stage of the proposed framework is done manually (i.e., the second stage is activated manually by the designer in each new scenario). However, it would be an interesting enhancement to add a component that can automatically detect a shift in the running scenario and activates the second stage.

\section{Experimental Design}
\label{sec:ExperimentalDesign}
In this section, we describe our experimental design for evaluating different aspects of the proposed method. The first experiment ascertains that the skill-sampling method we have used from the literature \cite{Dilokthanakul17,SukhbaatarKSF17} is working as intended. This is done by comparing to random skill-sampling. The second experiment compares our end-to-end system with two state-of-the-art methods from the MORL literature.

\subsection{Experimental Environment}
\begin{figure*}[tp]
   \begin{minipage}[]{.32\linewidth}
     \centering
     \includegraphics[width=5.5cm,height=4cm]{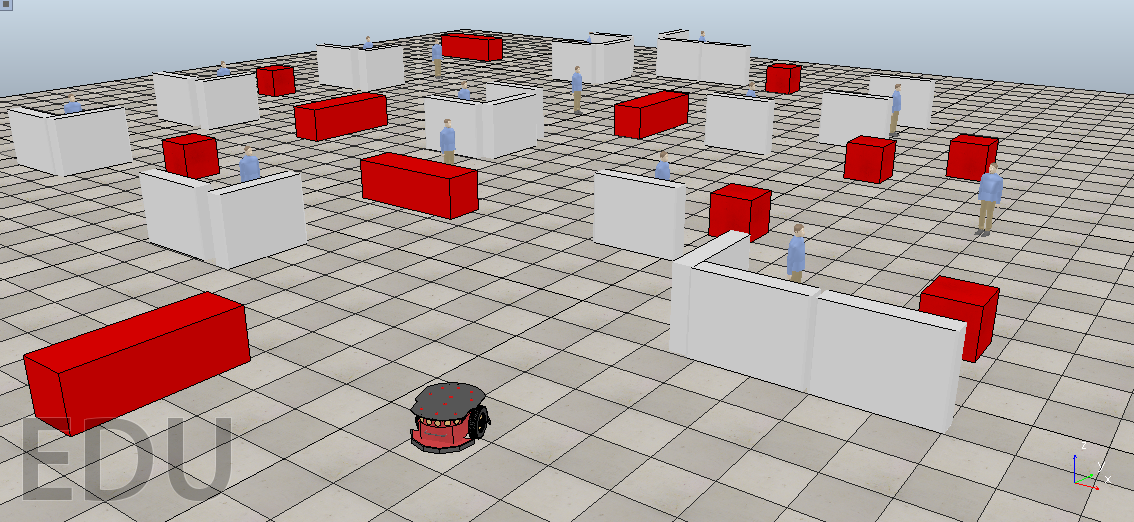}
     \subcaption{}\label{fig:SAR}
   \end{minipage}
   \begin{minipage}[]{.32\linewidth}
     \centering
     \includegraphics[width=5.5cm,height=4cm]{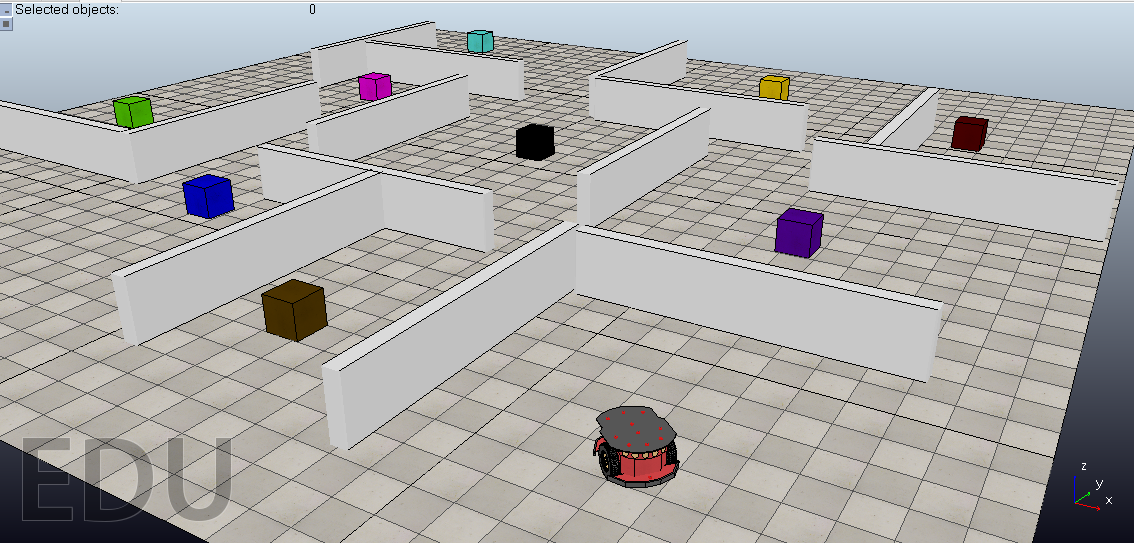}
     \subcaption{}\label{fig:DST}
   \end{minipage}
   \begin{minipage}[]{.32\linewidth}
     \centering
     \includegraphics[width=5.5cm,height=4cm]{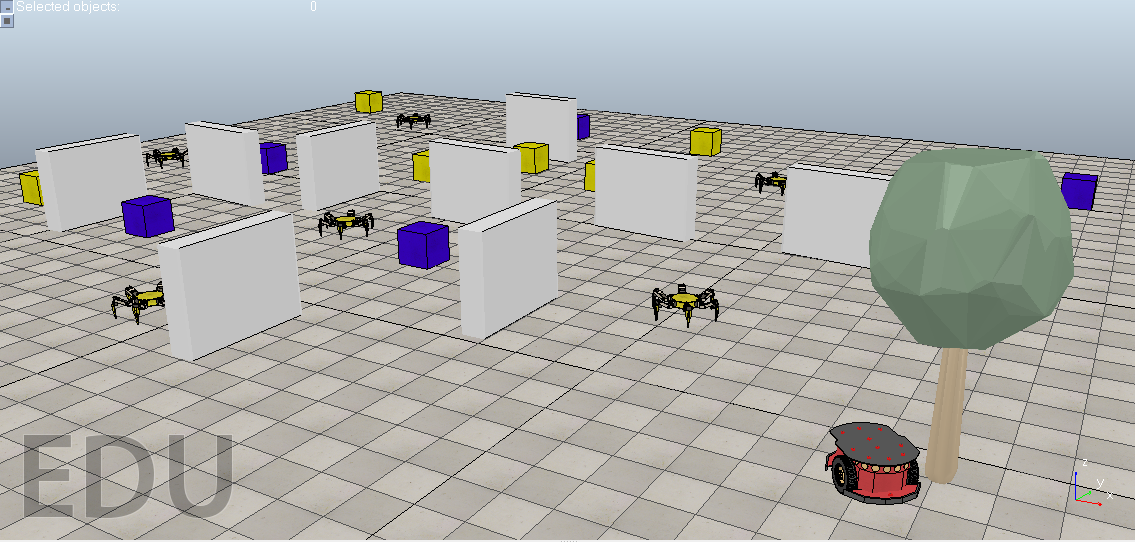}
     \subcaption{}\label{fig:RG}
   \end{minipage}
   \caption{Layouts of different scenarios representing non-stationary dynamics in the environment. (a) The search and rescue (SAR) scenario. Human oracles represent victims, walls represent obstacles, while red cubes represent fire danger. (b) The treasure Search (TS) Scenario. Different colored cubes represent different treasure values and walls represent obstacles. (c) The resource gathering (RG) scenario. Spider-like robots represent enemies, blue cubes represent gym resources, yellow cubes represent gold resources, walls represent obstacles, and the tree object represents the home location.}
   \label{fig:Envs}
\end{figure*}

In order to simulate real-world dynamic environments that demand a continuous learning process, we designed a complex robotics environment that consists of a static setup including the physical agent's build, the characteristics of the terrestrial surface, and the presence of obstacle objects. Additionally, dynamic setups can be introduced to the environment through three different scenarios that add new states to the state space and new reward functions.  The static setup simulates the basic operation scenario in which the agent can learn a set of generic skills. While each dynamic scenario represents new characteristics that can arise over time in the environment demanding more advanced skills from the learning agent. Therefore, we can experimentally evaluate the continuous learning ability of the proposed method in such complex and dynamic environments. The environment was implemented using the V-REP\footnote{\url{http://www.coppeliarobotics.com/}} robotics simulation environment \cite{VREP_Env}. The learning agent is simulated using the Pioneer 3-DX\footnote{\url{https://robots.ros.org/pioneer-3-dx/}} robot model defined in the V-REP environment. This robot has $16$ proximity sensors mounted around it in addition to a camera sensor mounted in the front. Accordingly, the state space is $S=\left\{ b^{1},b^{2},\ldots,b^{16},Cam\right\}$, where $\left\{ b^{1},b^{2},\ldots,b^{16}\right\}$ is the set of state features representing proximity sensors readings, and $\left\{Cam\right\}$ is the camera sensor reading represented by $64\times64$ frame flattened into a $4096$ array. The action space $A=\left\{ F^{left},F^{right},D^{left},D^{right}\right\}$  which includes the force on the left wheel motor, the force on the right wheel motor, rotation direction of the left wheel, and rotation direction of the right wheel. The state and action representations are the same among the three scenarios as we use the same learning agent. Figure \ref{fig:Envs} shows the layout of the robotic scenarios. These dynamic scenarios are based on well-established benchmarks in MORL literature \cite{Vamplew2011,Sherif_18} that were originally introduced as grid-worlds. In this paper, we introduce continuous variants of them that are more representative for real-world robotic scenarios. We provide the V-REP scene source code for these scenarios for possible future use\footnote{\url{https://figshare.com/collections/V-REP_MORL_Environments/4657700}}.

\setcounter{subfigure}{0}

\textbf{Search and Rescue (SAR) Scenario}

In this scenario, the agent gets a vector of three rewards $\vec{r}=\left[\,\mathit{r}^{\mathit{victim}},\,\mathit{r}^{\mathit{fire}},\right],\,\vec{r}\in\mathbb{R\mathrm{^{\mathit{2}}}}$. The victim reward function $r^{victim}$ is $+3$ for each detected victim and $0$ elsewhere, the fire penalty function $r^{fire}$ is $-5$ for each exposure and $0$ elsewhere. The added stochastic state transition is defined as random death probability of each human victim $\xi_{i},\,i\in\left\{ 1,2,3,\ldots,N\right\} $
for $N$ victims.  

\textbf{Treasure Search (TS) Scenario}

In this scenario, multiple treasures can be found, each
with a different reward value. There are two objectives. First, to minimize
needed time to find the treasure. Second, to maximize the treasure's
value. Accordingly, the reward vector has two rewards $\mathbf{\mathrm{\overrightarrow{r}=\left[\,\mathit{r}^{\mathit{time}},\,\mathit{r}^{\mathit{treasure}}\right],\,\mathbf{\mathrm{\vec{r}}\in\mathbb{R\mathrm{^{\mathit{2}}}}}}}$,
where $r^{time}$ is a time penalty of $-1$ on all turns and $r^{treasure}$
is the captured treasure reward which depends on the treasure's value.

\textbf{Resources Gathering (RG) Scenario}

This scenario simulates a task to collect resources (gold
and gems) and return home while avoiding attacks from enemy objects. The enemy attack may
occur with a $10\%$ probability if the learning agent is within the enemy's area. If an attack happens, the agent
loses any resources currently being carried and is returned to the
home location. The objectives are to maximize
the resources gathered while minimizing enemy attacks. The rewards
vector is defined as $\mathbf{\mathrm{\overrightarrow{r}=\left[\,\mathit{r}^{\mathit{resources}},\,\mathit{r}^{enemy}\right],\,\mathbf{\mathrm{\vec{r}}\in\mathbb{R\mathrm{^{\mathit{2}}}}}}}$,
with $r^{resources}$ is $1$ for each resource collected and $r^{enemy}$
is $-1$ for each attack.

\subsection{Experiments}

\subsubsection{Evaluating the Evolution of Generic Skill Sets in MOMDPs}~\\

\textbf{Objective: }The objective of this experiment is to confirm the capability of the IMRL skill sampling method \cite{Dilokthanakul17,SukhbaatarKSF17} in learning skills in the static part of the environment.

\textbf{Assumption: }Intrinsically motivated skill sampling can focus the learning process on skills that match the current performance level of the learning agent, therefore, it can achieve better skill coverage and overall performance.

\textbf{Comparison Algorithms: }We compare the IMRL sampler with a random sampler which uniformly samples skills from the predefined skill set. Thus, we can ensure the effectiveness of the IMRL sampling method. We refer to the used IMRL sampler as generic intrinsically motivated exploration (GIME) when reporting the results.

\textbf{Method: }We designed a generic set of useful goals (defined by reward functions) relevant to the learning agent’s physical structure and the static characteristics of the environment common to all three scenarios. There are ten defined goals. The move forward skill is rewarded when the agent moves the wheel forward in a stable manner, while its quick version rewards it proportionally with the achieved revolutions per minute (rpm) of the wheels in the forward direction. The move backward and its quick version are exact opposite of the move 
forward skills. The turn right skill is rewarded when the agent puts more force on the right wheel, while its quick version rewards it proportionally with the rpm to encourage faster movement. The turn left skills are exact opposite of the turn right skills. The bounce back skill is penalized when the agent has an object within $\tau$ distance on the proximity sensor. Finally, the turn-around skill is rewarded when the agent perceives a wall in a different part of its camera view over successive time-steps.

\begin{figure}
\begin{centering}
\includegraphics[width=9cm,height=5cm]{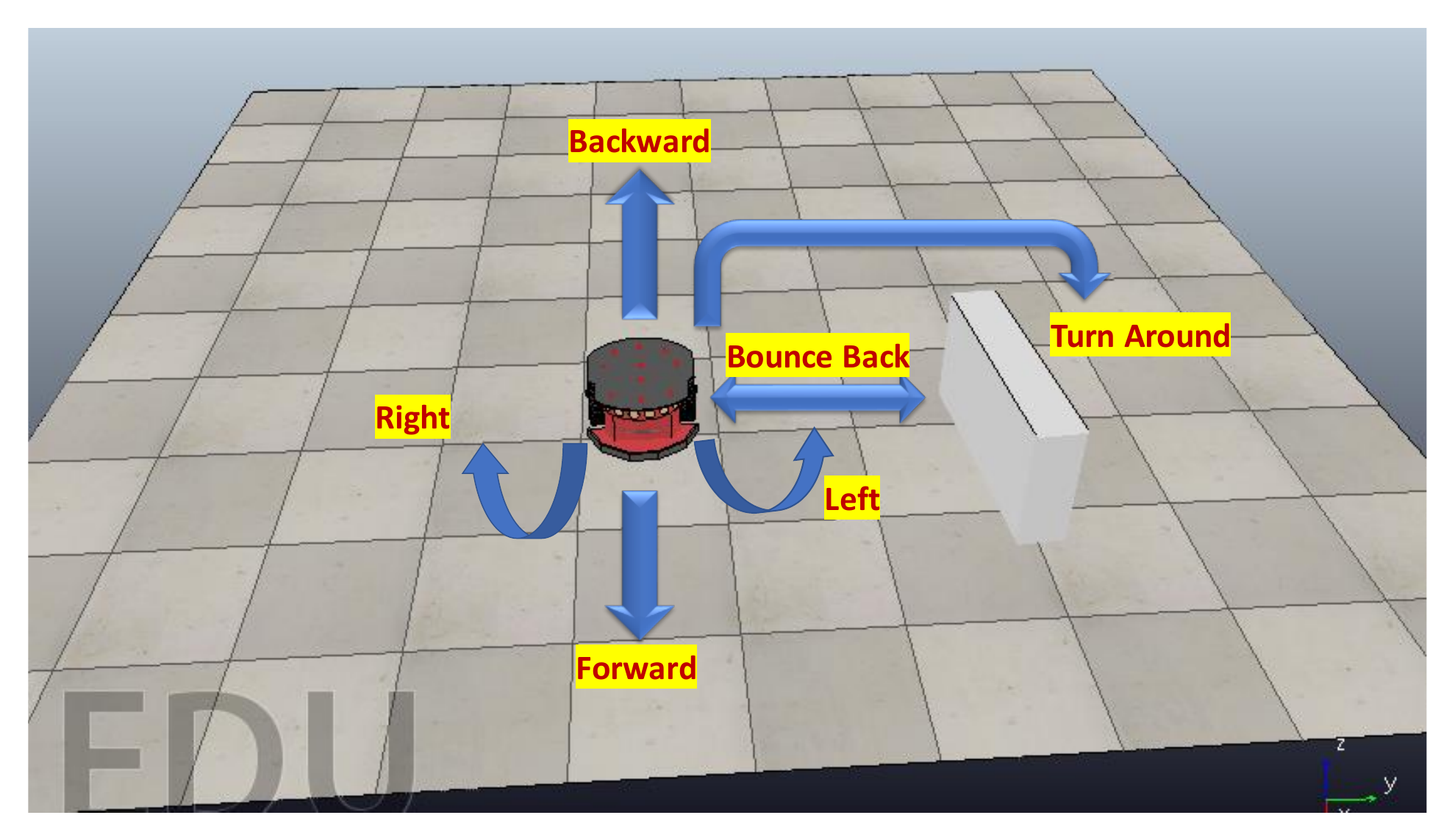}
\par\end{centering}
\caption{Visual illustration of the predefined skill set for the static environment setup.}

\label{fig:GenericSkillRobo}
\end{figure}

Table \ref{tbl:SkillRobotics} summarizes the predefined skill set. Figure \ref{fig:GenericSkillRobo} depicts a visual illustration for each skill in the static environment setup. Each method will sample from this predefined set during the evaluation. We conduct 15 independent runs. Each run includes 100 learning cycles. Each cycle includes 2500 episodes. We use a time-bounded episode configuration of 150 time-steps.  

\begin{table*}

\caption{Generic skill set and dedicated reward functions}

\begin{centering}
\begin{tabular}{|c|c|c|}
\hline 
\textbf{Skill} & \textbf{Description} & \textbf{Reward Function}\tabularnewline
\hline 
Move Forward & Advance in the forward direction & $r=\begin{cases}
1 & F^{right}\simeq F^{left}\,\&\,D^{right}=D^{left}=+1\\
0 & Otherwise
\end{cases}$\tabularnewline
\hline 
Move Backward & Advance in the backward direction & $r=\begin{cases}
1 & F^{right}\simeq F^{left}\,\&\,D^{right}=D^{left}=-1\\
0 & Otherwise
\end{cases}$\tabularnewline
\hline 
Turn Right & Turning to the right & $r=\begin{cases}
1 & F^{right}>F^{left}\,\&\,D^{right}=D^{left}=+1\\
0 & Otherwise
\end{cases}$\tabularnewline
\hline 
Turn Left & Turning to the left & $r=\begin{cases}
1 & F^{right}<F^{left}\,\&\,D^{right}=D^{left}=+1\\
0 & Otherwise
\end{cases}$\tabularnewline
\hline 
Quick Forward & Quick advance in the forward direction & $r=\begin{cases}
rpm & F^{right}\simeq F^{left}\,\&\,D^{right}=D^{left}=+1\\
0 & Otherwise
\end{cases}$\tabularnewline
\hline 
Quick Backward & Quick advance in the backward direction & $r=\begin{cases}
rpm & F^{right}\simeq F^{left}\,\&\,D^{right}=D^{left}=-1\\
0 & Otherwise
\end{cases}$\tabularnewline
\hline 
Quick Turn Right & Quick turning to the right & $r=\begin{cases}
rpm & F^{right}>F^{left}\,\&\,D^{right}=D^{left}=+1\\
0 & Otherwise
\end{cases}$\tabularnewline
\hline 
Quick Turn Left & Quick turning to the left & $r=\begin{cases}
rpm & F^{right}<F^{left}\,\&\,D^{right}=D^{left}=+1\\
0 & Otherwise
\end{cases}$\tabularnewline
\hline 
Bounce Back & Move away from an obstacle & $r=\begin{cases}
-1 & Any(b)<\tau\\
0 & Otherwise
\end{cases}$\tabularnewline
\hline 
Turn Around & Get to the opposite side of an obstacle & $r=\begin{cases}
1 & obstacle\,in\,other\,direction\,of\,Cam\,view\\
0 & Otherwise
\end{cases}$\tabularnewline
\hline 
\end{tabular}
\par\end{centering}
\label{tbl:SkillRobotics}
\end{table*}

\textbf{Evaluation Criteria: } We use two performance metrics in this experiment. First, is the average success rate for each skill over the last 10 learning cycles. Second, is the average skill exploration rate. We show the average and standard deviation for each of these metrics over the 15 conducted runs.

\subsubsection{Assessing the Impact of Hierarchical Policy Learning in MOMDPs}~\\

\textbf{Objective: }In this experiment, we aim at assessing the impact of learning hierarchical policies using the evolved generic skill set to generalize over different shifts in the environment's dynamics (i.e. scenarios).
 
\textbf{Assumption: }Our assumption is that bootstrapping the learning process with a generic set of skills can enhance the learning of specialized policies in each different scenario.
	
\textbf{Comparison Algorithms: }We compare our proposed method with three other algorithms. The first two are widely considered state-of-the-art algorithms in MORL literature \cite{roijers2013survey} including the optimistic linear support (OLS) algorithm \cite{Roijers2014}, and the threshold lexicographic ordering (TLO) algorithm \cite{geibel2006reinforcement}. While the third is our previously proposed intrinsically motivated MORL method named IM-MORL during the comparison \cite{Sherif_18}. The IM-MORL method uses intrinsic motivation for preference exploration in evolving robust policy coverage sets that can solve the MOMDP problem. However, this method does not have a generic skill learning stage and it directly learns the policy coverage sets for a specific scenario, therefore, it is limited in generalizing to different scenarios. Finally, the proposed methodology in this paper is referred to as generic intrinsically motivated MORL (GIM-MORL) during the comparison. 
    
\textbf{Method: }We conduct this experiment in two scenario groups: stationary scenarios; and non-stationary scenarios. In the former, the distribution of objects is stationary in each run. In the latter, the distribution of objects is non-stationary, as $25$\% of them change their locations randomly every $100$ episodes. For each group, we execute $15$ runs that differ in the initial distribution of objects. Each run is divided into a training phase and a testing phase, each of $2500$ episodes. The training phase allows each method to evolve its policy coverage set. While in the testing phase, we sample ten user preferences uniformly (see Table \ref{tbl:U}), and every $250$ episodes the preference changes to evaluate the performance of the evolved policy coverage set for each method. For the parameter configuration of the OLS, TLO, and RFPB algorithms we follow the
same configuration in \cite{Roijers2014}, \cite{geibel2006reinforcement}, and \cite{Sherif_18}, respectively.

\begin{table*}
\caption{The set of uniformly sampled user preferences used in
the experimental design.}
\begin{centering}
\begin{tabular}{c|cccccccccc}
\hline 
\multirow{1}{*}{Preference} &  \multicolumn{1}{c|}{$P_{1}$} & \multicolumn{1}{c|}{$P_{2}$} & \multicolumn{1}{c|}{$P_{3}$} & \multicolumn{1}{c|}{$P_{4}$} & \multicolumn{1}{c|}{$P_{5}$} & \multicolumn{1}{c|}{$P_{6}$} & \multicolumn{1}{c|}{$P_{7}$} & \multicolumn{1}{c|}{$P_{8}$} &
 \multicolumn{1}{c|}{$P_{9}$} & \multicolumn{1}{c}{$P_{10}$} 
\tabularnewline
\hline 
\multicolumn{1}{c}{$w_{1}$} & 0.66 & 0.33 & 0.28 & 0.54 & 0.68 & 0.44 & 0.88 & 0.65 & 0.48 & 0.71\tabularnewline
\hline 
\multicolumn{1}{c}{$w_{2}$} & 0.34 & 0.67 & 0.72 & 0.46 & 0.32 & 0.56 & 0.12 & 0.35 & 0.52 & 0.29 \tabularnewline
\hline 
\end{tabular}
\par\end{centering}
\label{tbl:U}
\end{table*}
   
\begin{figure}
\begin{centering}
\includegraphics[width=7cm, height=6cm]{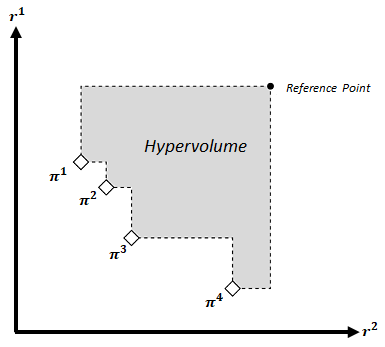}
\par\end{centering}
\caption{An example for a graphical representation of the \textit{hypervolume} metric in a two dimensional reward space. In this example, the metric is represented by the area of the shaded shape bounded by the policies in the policy coverage set and the reference point. }
\label{fig:hypermetric}

\end{figure}

\textbf{Evaluation Criteria: }We evaluate the three comparative methods over two main metrics. First, the sum of median rewards metric, which is calculated by taking the median reward value for each preference, sum them for each run, then taking the average of this sum over the 15 runs. This metric reflects the overall performance of the evolved policy coverage set for each algorithm over the 15 independent runs executed. For visualizing this evaluation, we show the average median value with standard deviation for each sampled preference. Second, the \textit{hypervolume}, which represents the volume of the space dominated by policy points from the policy coverage set in the reward space given a reference point. This metric measures the quality of the evolved policy coverage set by a multi-objective optimization algorithm \cite{HyperVolume_2009}. Figure~\ref{fig:hypermetric} shows an example of the \textit{hypervolume} in a two dimensional reward space. The higher the value of this metric the better the policy coverage set. We followed the algorithm described in \cite{HyperVolume_2009} to calculate the value of this metric. This algorithm works in an iterative manner to approximate the hypervolume area as the sum of rectangular areas bounded by the reference point and each policy point in the policy coverage set.

\section{Results and Discussion} \label{sec:Results}
In this section, we present and discuss the results of the conducted experiments.

\subsection{Evolving Generic Skill Sets}

\begin{figure}
\begin{centering}
\includegraphics[width=9cm,height=7.5cm]{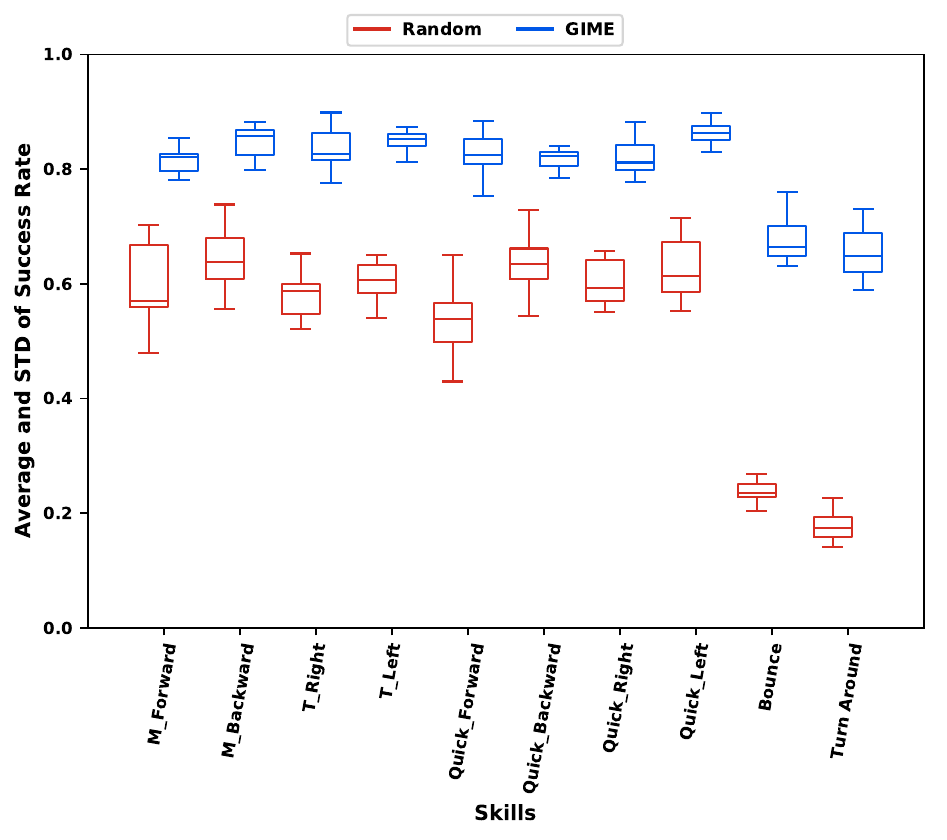}
\par\end{centering}
\caption{Boxplots for skill success ratio for each method over 15 runs
in the static environment setup.}
\label{fig:Exp1_robo_success}
\end{figure}

Results for the average skill success ratio metric in the static environment's setup are depicted in Figure \ref{fig:Exp1_robo_success} through boxplots for each skill. It can be noticed that GIME significantly outperformed the random exploration method in all skills and this performance margin is magnified in the bounce back and turn around skills. The reason for this finding lies in the ability of GIME to target skills that have unpredictable performance (i.e., unstable policy), therefore it facilitates focusing the learning process on those skills that need enhancements. This ability is further confirmed in the performance results of the bounce back and turn around skills which are the most difficult ones in comparison to the rest. While the random exploration method does not have any clue on the current performance level of the learning agent with respect to each skill as it samples goals uniformly from the predefined skill set.  

\begin{figure}
\begin{centering}
\includegraphics[width=9cm,height=7.5cm]{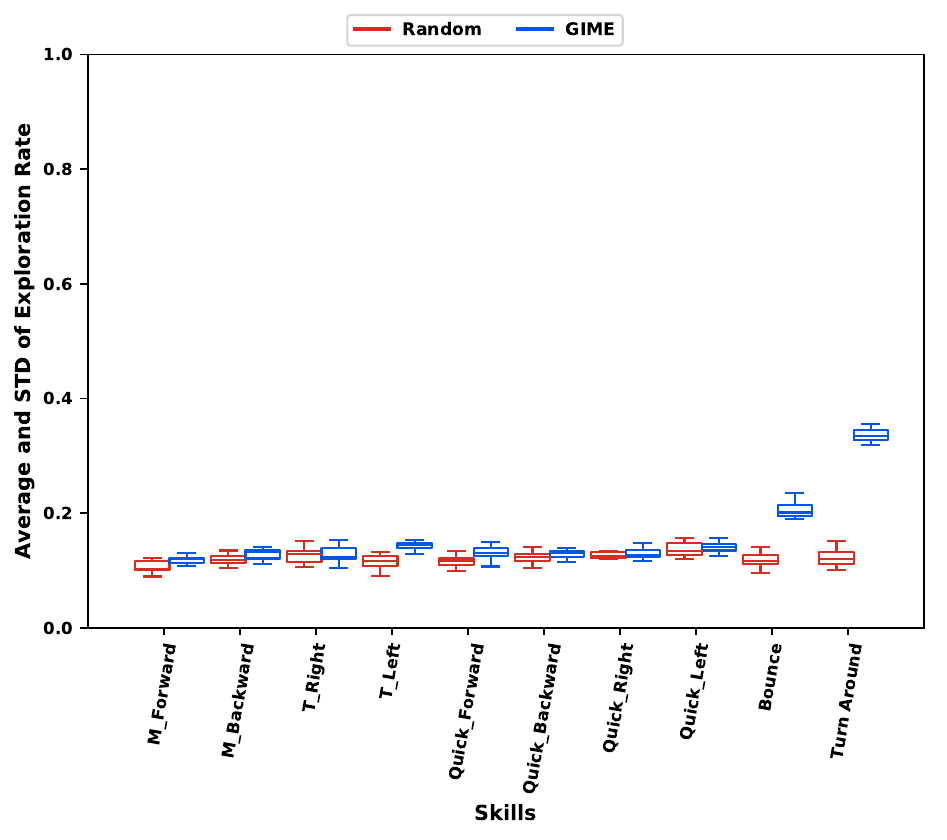}
\par\end{centering}
\caption{Boxplots for skill exploration ratio for each method over 15 runs
in the static environment setup.}

\label{fig:Exp1_robo_exp}
\end{figure}

The second evaluation metric helps to further understand the behavior of each goal exploration method. The skill exploration ratio measures the number of times a specific skill was proposed to the learning agent out of the total number of learning cycles. Figure \ref{fig:Exp1_robo_exp} presents the average skill exploration ratio through boxplots for each method in the environment. It can be noticed that the random exploration method has a consistent and closely equal exploration ratio overall skills, which confirms its working mechanism of uniformly sampling skills from the predefined set. The results for GIME are heavily skewed around the `bounce back' and `turn around' skills, as these skills are the most challenging ones in comparison to the rest in the predefined set. Therefore, the GIME method increases the exploration rate for these skills in order to reach stable policies which have predictable performance. This finding confirms the adaptability of the exploration behavior of the IMRL method in comparison to the fixed behavior of the random exploration method.

\subsection{Evolving Policy Coverage Sets for MOMDP Tasks}

In this section, we present the comparison results between the proposed method (GIM-MORL) and the three other comparative methods over both stationary and non-stationary scenarios.
 
\subsubsection{Performance Comparison in Stationary Scenarios}

\begin{figure}
   \begin{minipage}[]{1\linewidth}
     \centering
     \includegraphics[width=9cm,height=6.5cm]{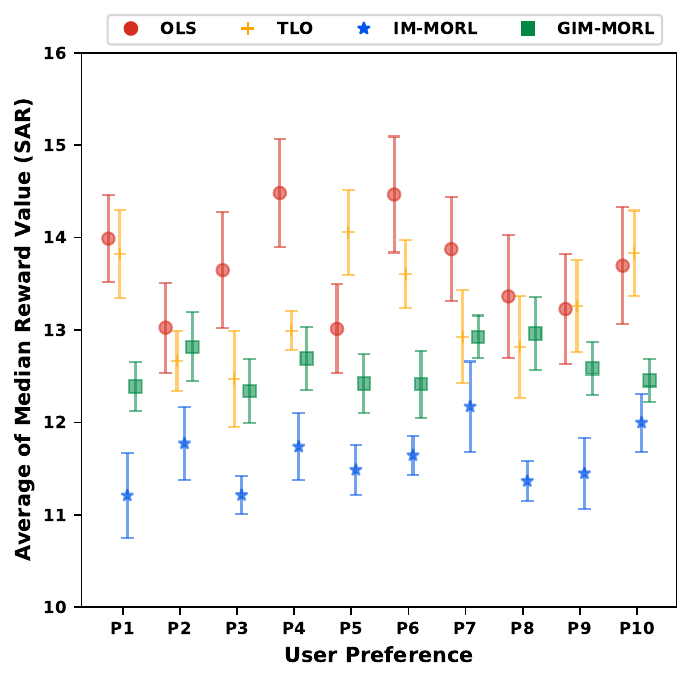}
     \subcaption{}\label{fig:exp_SAR}
   \end{minipage}
   \begin{minipage}[]{1\linewidth}
     \centering
     \includegraphics[width=9cm,height=6.5cm]{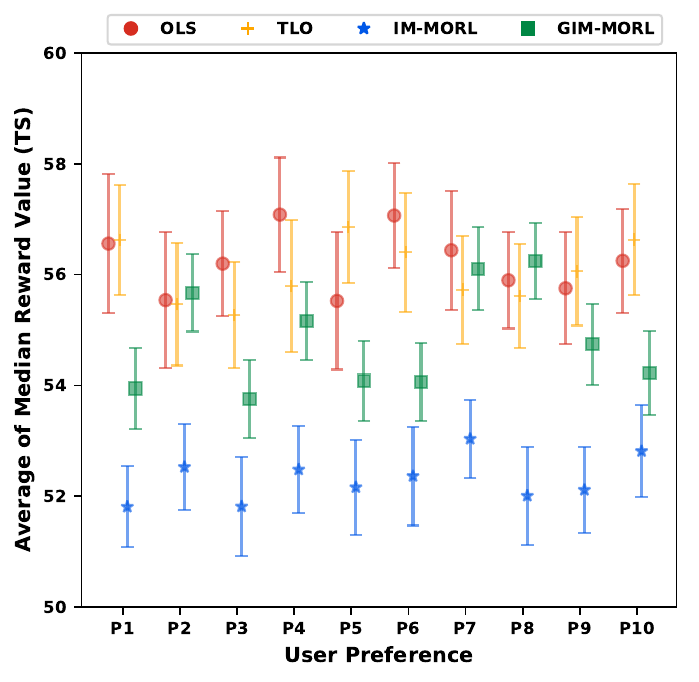}
     \subcaption{}\label{fig:exp_DST}
   \end{minipage}
   \begin{minipage}[]{1\linewidth}
     \centering
     \includegraphics[width=9cm,height=6.5cm]{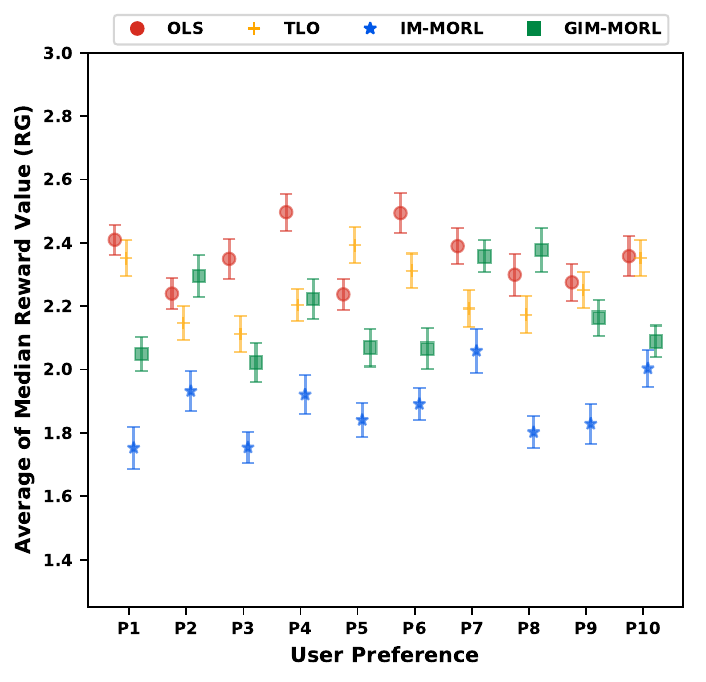}
     \subcaption{}\label{fig:exp_RG}
   \end{minipage}
   \caption{ Results for the median reward metric in stationary scenarios experiment averaged over 15 independent runs with standard deviation bars. (a) The search and rescue (SAR) scenario. (b) The treasure search (TS) scenario. (c) The resource gathering (RG) scenario.}
   \label{fig:Exp_Stationary}
\end{figure}

Recalling from our experimental design, the stationary scenario experiment targets evaluating the performance of the methods with the assumption of stationary state space dynamics. Under this assumption, the parameters of the state transition distribution ($T\,:\,S\times A\times S^{\prime}\rightarrow\left[0,1\right]$) are the same between training and testing sessions for each method. Figure \ref{fig:Exp_Stationary} shows the results of the average median reward metric with standard deviation bars over the 15 independent runs conducted. 

\setcounter{subfigure}{0}

While the OLS and TLO methods achieved the highest results, the IM-MORL and GIM-MORL methods achieved comparable results. This difference in result is due to targeting specialized results by the former comprehensive search methods in contrast to the latter methods which target robust steppingstone policies instead. Moreover, it can be noticed that the proposed method (GIM-MORL) outperformed our previously proposed method (IM-MORL) due to the use of the learned generic skill set in the previous stage, which proved to enhance the performance of the resultant policies.

\begin{figure}
\begin{centering}
\includegraphics[width=9cm,height=6.5cm]{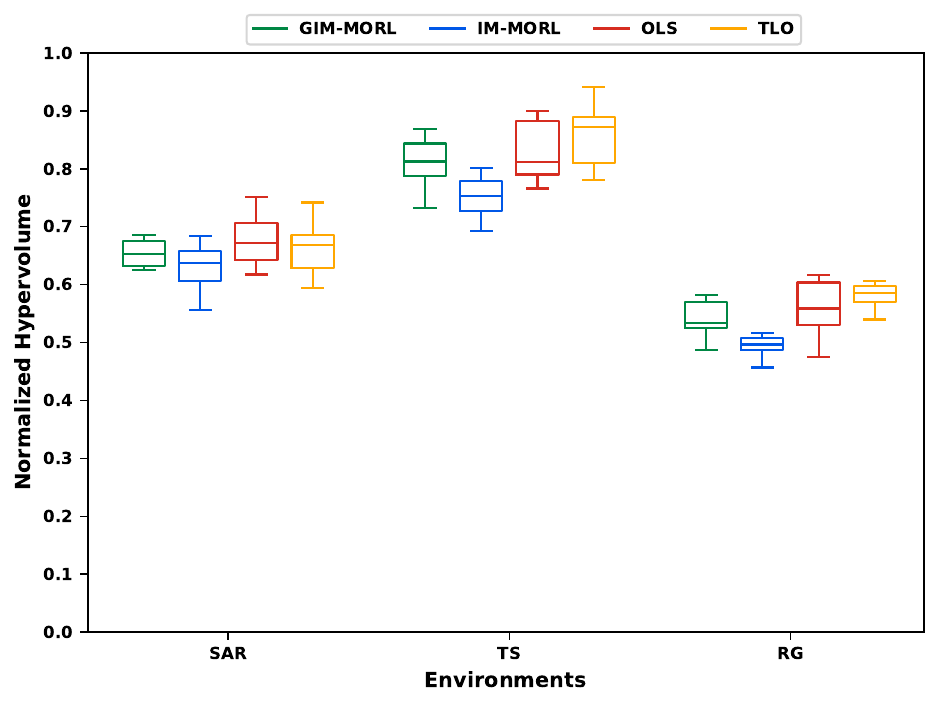}
\par\end{centering}
\caption{Boxplots for the normalized \textit{hypervolume}
metric in stationary environments averaged over 15 independent runs for each algorithm. Results are grouped by each environment.}
\label{fig:hv_sta}
\end{figure}

For the results on the second evaluation metric, Figure \ref{fig:hv_sta} presents boxplots for normalized hypervolume values for each algorithm over the 15 independent runs conducted. These results confirm the finding from the previous evaluation metric as the OLS and the TLO methods achieved the higher results followed by the GIM-MORL and the IM-MORL methods. In addition, the normalized hypervolume metric better reflects the comparable quality of the resultant policy coverage sets by the latter methods, which was close to those of the former methods on average.

It has to be noted that the proposed GIM-MORL method was able to achieve these results on three different scenarios by utilizing the same generic skill set learned in the previous stage of the methodology. Based on these findings, it is noticeable that learning a generic skill set can help in enhancing the performance of the MORL method and gives it the ability to generalize on different scenarios.
   
\subsubsection{Performance Comparison in Non-Stationary Scenarios}

We present the method comparison results in non-stationary scenarios. In such environments, the parameters of the state transition distribution ($T\,:\,S\times A\times S^{\prime}\rightarrow\left[0,1\right]$) can vary over time (i.e. entities in the environment are allowed to change their location over time). Such scenarios are more common in real-world applications, therefore, we target in this part of the experimental design to analyze the impact of such scenarios on the performance of each method. 

\begin{figure}
\begin{centering}
\includegraphics[width=9cm,height=6.5cm]{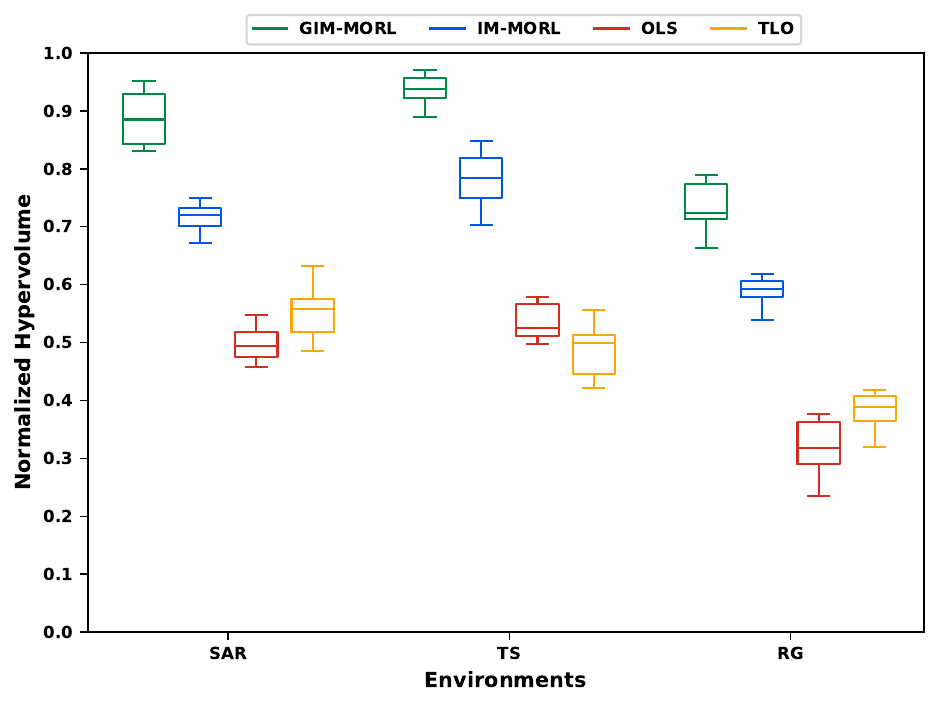}
\par\end{centering}
\caption{Boxplots for the normalized \textit{hypervolume}
metric in non-stationary environments averaged over 15 independent runs for each algorithm. Results are grouped by each environment.}
\label{fig:hv_nonsta}
\end{figure}

\begin{figure}
   \begin{minipage}[]{1\linewidth}
     \centering
     \includegraphics[width=9cm,height=6.5cm]{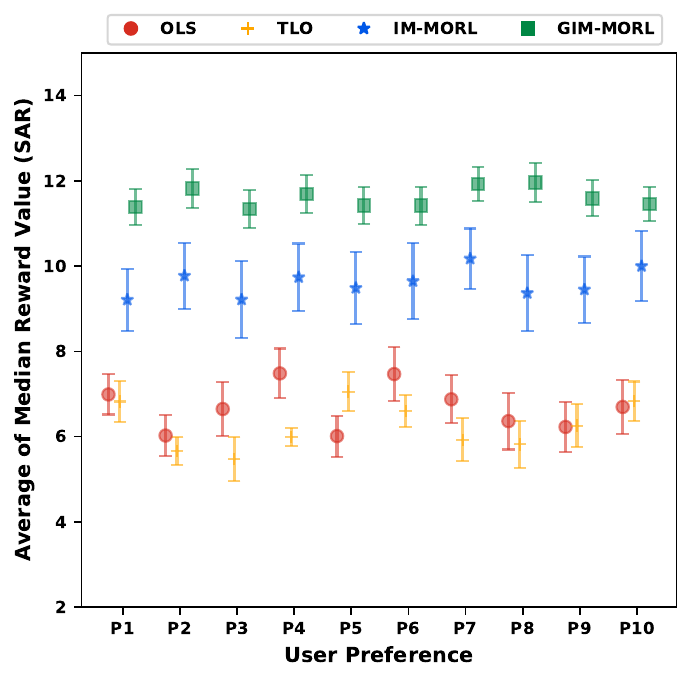}
     \subcaption{}\label{fig:exp2_SAR}
   \end{minipage}
   \begin{minipage}[]{1\linewidth}
     \centering
     \includegraphics[width=9cm,height=6.5cm]{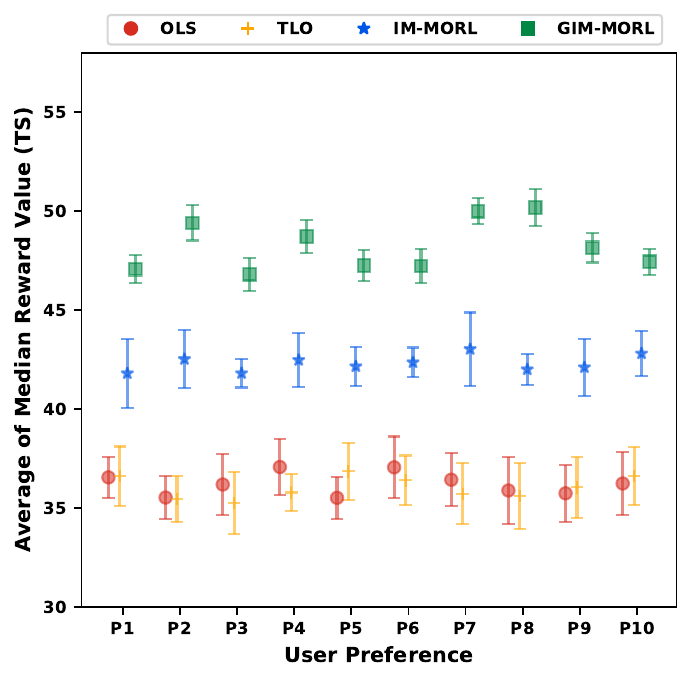}
     \subcaption{}\label{fig:exp2_DST}
   \end{minipage}
   \begin{minipage}[]{1\linewidth}
     \centering
     \includegraphics[width=9cm,height=6.5cm]{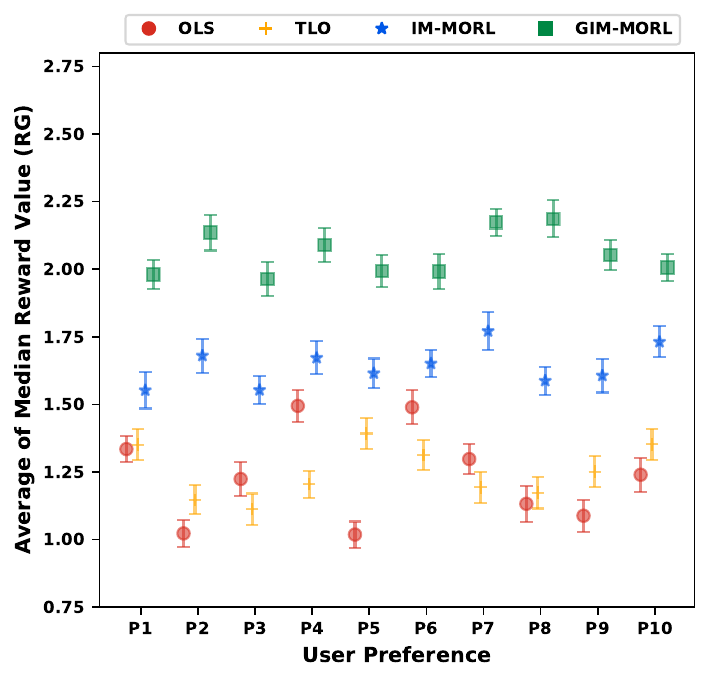}
     \subcaption{}\label{fig:exp2_RG}
   \end{minipage}
   \caption{Results for the median reward metric in non-stationary scenarios experiment averaged over 15 independent runs with standard deviation bars. (a) The search and rescue (SAR) scenario. (b) The treasure search (TS) scenario. (c) The resource gathering (RG) scenario.}
   \label{fig:Exp_NonStationary}
\end{figure}

Figure \ref{fig:Exp_NonStationary} depicts the results for the average median reward value with standard deviation bars over the 15 independent runs executed. It is observable that the OLS and TLO methods have significant degradation in their results in comparison to the stationary scenarios experiment. Mainly, this is due to their assumption of a stationary state transition distribution and the targeting greedy specialized policies that were tailored for a specific environment setup that went outdated with the ongoing dynamics in the state space. In contrast, the IM-MORL and GIM-MORL methods did not suffer from such significant performance degradation due to their ability to adaptively revisit the affected preference regions after drifts in the state space dynamics and targeting robust steppingstone policies that are optimized for performance stability over wider preference intervals. Moreover, the proposed GIM-MORL significantly (p-value < $0.05$) outperformed the other methods over the three scenarios on this evaluation metric with average percentages of $57\%$, $30\%$, and $65\%$ for SAR, TS, and RG, respectively.

\setcounter{subfigure}{0}

The normalized hypervolume results presented in Figure \ref{fig:hv_nonsta} confirmed the previous finding as the quality of the resultant policy coverage sets for the OLS and TLO methods significantly degraded in comparison to the stationary scenarios experiment. While the quality of the sets generated by IM-MORL and GIM-MORL methods was not significantly impacted in a similar manner. The proposed GIM-MORL method significantly outperformed (p-value < $0.05$) the other methods with an average percentage of $50\%$ over the three scenarios on this metric. This emphasizes the performance gain achieved through learning generic skills and utilizing them to generalize on different scenarios. 

\section{Conclusion}
\label{sec:conc}
In this paper, we proposed a novel intrinsically motivated multi-objective reinforcement learning method that can learn hierarchical policy coverage sets to better generalize to different shifts in the environment's dynamics including time-varying changes in the parameters of the state transition probability distribution and in the parameters of the reward functions prior distribution. We experimentally evaluated the performance of the generic skill learning and the evolution of hierarchical policy coverage sets for a complex robotics environment with non-stationary dynamics in the state and reward spaces. We compared our method with three state-of-the-art multi-objective reinforcement learning methods. Results showed that our proposed method generalized better to different scenarios, which enabled it to significantly outperform the other comparatives on the evaluation metrics.

In future work, we are going to explore techniques for automatically generating possible goals in a given scenario through intrinsic motivation exploration. In addition, we will investigate ways to prioritize and schedule skill learning based on relevancy aspects such as similar actions or goals, therefore we can speed up the learning process. Also, we are going to look for concept drift detection methods that can identify state space shifts across scenarios and automatically activates the second stage of the proposed GIM-MORL framework.


\ifCLASSOPTIONcaptionsoff
  \newpage
\fi

\bibliographystyle{plain}
\bibliography{FinalDB}

\end{document}